\documentclass[sigconf]{acmart}

\usepackage{balance}

\def\BibTeX{{\rm B\kern-.05em{\sc i\kern-.025em b}\kern-.08emT\kern-.1667em\lower.7ex\hbox{E}\kern-.125emX}}

\usepackage{algorithm}
\usepackage[noend]{algpseudocode}
\usepackage[utf8]{inputenc}

\algblockdefx{MRepeat}{EndRepeat}{\textbf{repeat}}{}
\algnotext{EndRepeat}

\algnewcommand\BoolAnd{\textbf{and}}
\algnewcommand\BoolOr{\textbf{or}}

\AtBeginDocument{%
  \providecommand\BibTeX{{%
    \normalfont B\kern-0.5em{\scshape i\kern-0.25em b}\kern-0.8em\TeX}}}

\setcopyright{acmlicensed}
\begin{document}

\fancyhead{}

\title{Hierarchical Affordance Discovery using Intrinsic Motivation}

\author{Alexandre Manoury}
\email{alexandre.manoury@imt-atlantique.fr}
\affiliation{%
  \institution{IMT Atlantique}
  \city{Brest}
  \state{France}
}


\author{Sao Mai Nguyen}
\email{mai.nguyen@imt-atlantique.fr}
\affiliation{%
  \institution{IMT Atlantique}
  \city{Brest}
  \state{France}
}

\author{C\'edric Buche}
\email{buche@enib.fr}
\affiliation{%
  \institution{ENIB}
  \city{Brest}
  \state{France}
}


\begin{abstract}
To be capable of life-long learning in a real-life environment, robots have to tackle multiple challenges. Being able to relate physical properties they may observe in their environment to possible interactions they may have is one of them. This skill, named affordance learning, is strongly related to embodiment and is mastered through each person's development: each individual learns affordances differently through their own interactions with their surroundings. Current methods for affordance learning usually use either fixed actions to learn these affordances or focus on static setups involving a robotic arm to be operated.

In this article, we propose an algorithm using intrinsic motivation to guide the learning of affordances for a mobile robot. This algorithm is capable to autonomously discover, learn and adapt interrelated affordances without pre-programmed actions. Once learned, these affordances may be used by the algorithm to plan sequences of actions in order to perform tasks of various difficulties. We then present one experiment and analyse our system before comparing it with other approaches from reinforcement learning and affordance learning.
\end{abstract}

\copyrightyear{2019}
\acmYear{2019}
\acmConference[HAI '19]{Proceedings of the 7th International Conference on Human-Agent Interaction}{October 6--10, 2019}{Kyoto, Japan}
\acmBooktitle{Proceedings of the 7th International Conference on Human-Agent Interaction (HAI '19), October 6--10, 2019, Kyoto, Japan}
\acmPrice{15.00}
\acmDOI{10.1145/3349537.3351898}
\acmISBN{978-1-4503-6922-0/19/10}

\thanks{The research work presented is partially supported by the European Regional Fund (FEDER) via the VITAAL Contrat Plan Etat Region}




\maketitle
\thispagestyle{fancy}
\lhead{}
\chead{
\texttt{\footnotesize{ A. Manoury, S. M. Nguyen, and C. Buche. Hierarchical Affordance Discovery using Intrinsic Motivation. In Human Agent Interaction, 2019.
 }}
\vspace{20pt}}
\rhead{}
\cfoot{}

\section{Introduction}

Continuous adaptation to the environment constitutes a key feature of the human learning process. It enables humans to learn to interact with newly discovered objects, either by reusing and adapting previously acquired knowledge or by building new skills more adapted to the situation at hand. This competence, named life-long learning, is one of the central challenges for service robots to act in our every day environment, towards socially assistive robotics and human agent interaction.

To tackle this, we adopt the approach of developmental robotics. Indeed, studying how infants learn and adapt constitutes an essential example of life-long learning. It highlights multiple mechanisms involved in this learning. Among them, we decide to focus on two in particular: the way infants relate what they see to how they may interact with surrounding objects; and how they explore and interact with their environment while building new skills.

The first one, coined since 1979 by Gibson as the concept of affordance \cite{Gibson79}, describes the strong relationship between visual cues and possible interactions. Contrarily to classical computer vision, central to the notion of affordance are the concepts of embodiment and of motor capabilities \cite{jamone_ugur2016_affordance_survey}. For instance, adults and infants do not see the same affordances for the same objects because they do not have the same body, the same way humans do not perceive the same affordances as robots. Such affordances evolve all along the life of a person, directly through its interactions with its surroundings.

The latter, the capacity that infants have to autonomously explore their environment, may be one of the answers of how affordances are learned. Indeed, infants use their curiosity to drive their exploration and build new skills through it \cite{Gottlieb2013}. This capacity, described as intrinsic motivation in psychology \cite{Miller1988}, provides a powerful mechanism to learn motor skills such as affordances.

In this paper, we focus on combining those two aspects of the human learning process: affordances and intrinsic motivation, by proposing a robotic framework to learn affordances thanks to active learning. We apply it to the case of a mobile robot. Moreover, as we aim at complex affordances that might require not only primitive actions but a succession of actions to be combined, we use planning to chain actions in order to perform task of various difficulty.


\section{Related work}

In this section we review the work related to two aspects of our approach: affordances learning and active learning algorithms, especially those using intrinsic motivation. We also present reinforcement learning methods we compare with later in the article.

\subsection{Affordances learning}

Many approaches to affordance learning has been developed: the traversability affordance for instance, has been studied in different works \cite{Ugur2010, 1641763}. Likewise, the grasp affordance is a recurrent topic and various approaches exist to learn it such as learning based on visual descriptors or raw image input \cite{5175529, Sergey2018}. However such methods are not easily generalised and tend to focus on one, or a fixed number of specific affordances, with no mechanism adapting it to new or more complex affordances.

More general approaches, not focusing only on one affordance, have also been proposed. Such approaches build and update a list of affordances through the robot interactions with its environment. In \cite{montesano_lopes2008_affordance_learning_coordination_imitation}, Bayesian Networks are used to learn dependencies between actions, effects, and visual properties; a strongly spread definition of affordances in robotics. Likewise, \cite{garcia_vayrac2016_affordance_learning} also uses a Bayesian approach but coupled with a fixed and finite pre-programmed action set to learn affordances. In our case, we aim to continual learning of multiple affordances through the interaction with its environment. Thus, the robot builds itself sensory motor skills using a wide variety of actions. The robot can use actions of unbounded length and duration, in a continuous action space. 
In \cite{ugur_nagai2015_affordance_imitation_motionese_staged_learning}, Ugur et al. propose a developmental approach of affordance learning. The robot learns by stages:  simple affordances first and more complex later. But they limit their approach to simple affordances with no multi-object interaction possible. In all of those methods above, the approach is limited to a single scene and to a single object interaction at a time, guiding the robot exploration without letting it choose autonomously. Furthermore, the considered setups usually focus on fixed robot with an end-point manipulator, while in our case we decide to consider a mobile robot, using its mobility to explore its surroundings by itself and choosing which object to interact with.

\subsection{Active motor learning}

Central to Gibson’s theory is the notion that the motor capabilities of the agent dramatically influence perception. Therefore affordance is closely linked to motor learning. In recent years, multiple approaches have emerged for active motor learning. For instance, several methods exist to learn forward and inverse models, map motor policies to sensorimotor outcomes; as formalised by \cite{FRANCIS1976457, WOLPERT19981317}. However, as the dimensionality of the spaces considered increase, the learner faces the curse of dimensionality \cite{Bellman1957} and no comprehensive method is possible.

Developmental methods, inspired by how infants explore and learn, have also been proposed to tackle this issue \cite{oudeyer_kaplan2004_iac}. Indeed, curiosity has been identified as a key mechanism for exploration \cite{Miller1988}. 
Other methods, even further inspired by human psychology has been developed \cite{baranes_oudeyer2009_riac}, using goal-babbling mechanism to generate goals and drive their exploration.

More recently, methods using intrinsic motivation to build a hierarchy of interrelated skills has been proposed. Firstly by using a pre-programmed and static hierarchy \cite{Forestier2016}, then by learning it through exploration for robotic arms \cite{Duminy2019FN} or mobile robots \cite{M2019}. The latter provides an algorithm, CHIME, that uses intrinsic motivation to guide its exploration. It is capable of adaptive addition and modification of a skill hierarchy based on its interactions. It may also plan of sequence of actions to perform complex tasks.

\subsection{Reinforcement learning}

In other domains, such as reinforcement learning, numerous methods have emerged to tackle solving daily tasks, using either classical approaches, such as Q-Learning or more recent neural network based ones: DQN\cite{mnih2015humanlevel} or Actor Critic algorithm\cite{MnihBMGLHSK16}. But such methods differ from our approach as they are designed to tackle specific tasks, defined by a reward function that need to be provided to the learner. For a more general approach of reinforcement learning methods, Universal Value Function Approximators \cite{Schaul2015} have been proposed: the task goal is this time learned as an environmental state instead of being fixed. This lets the learned value function to be more general and applicable to various goals. Closer to our approach, the CURIOUS algorithm has been proposed, combining deep reinforcement learning and intrinsic motivation to generate goals to explore  \cite{Colas2019P3ICML}.

We decide to base our proposition on the CHIME algorithm and adapt their learning algorithm to the affordance learning problem by using sensorimotor features. Whereas in \cite{M2019}, CHIME could not generalise its skills to new objects, our algorithm is capable to generalise to new objects, as its learning is based on sensory features.

\section{Proposition}

Before describing our method, we first present the experiment and formalise the learning problem. 

\subsection{Setup} \label{section:setup}

\begin{figure}[h!]
    \centering
    \includegraphics[width=\linewidth]{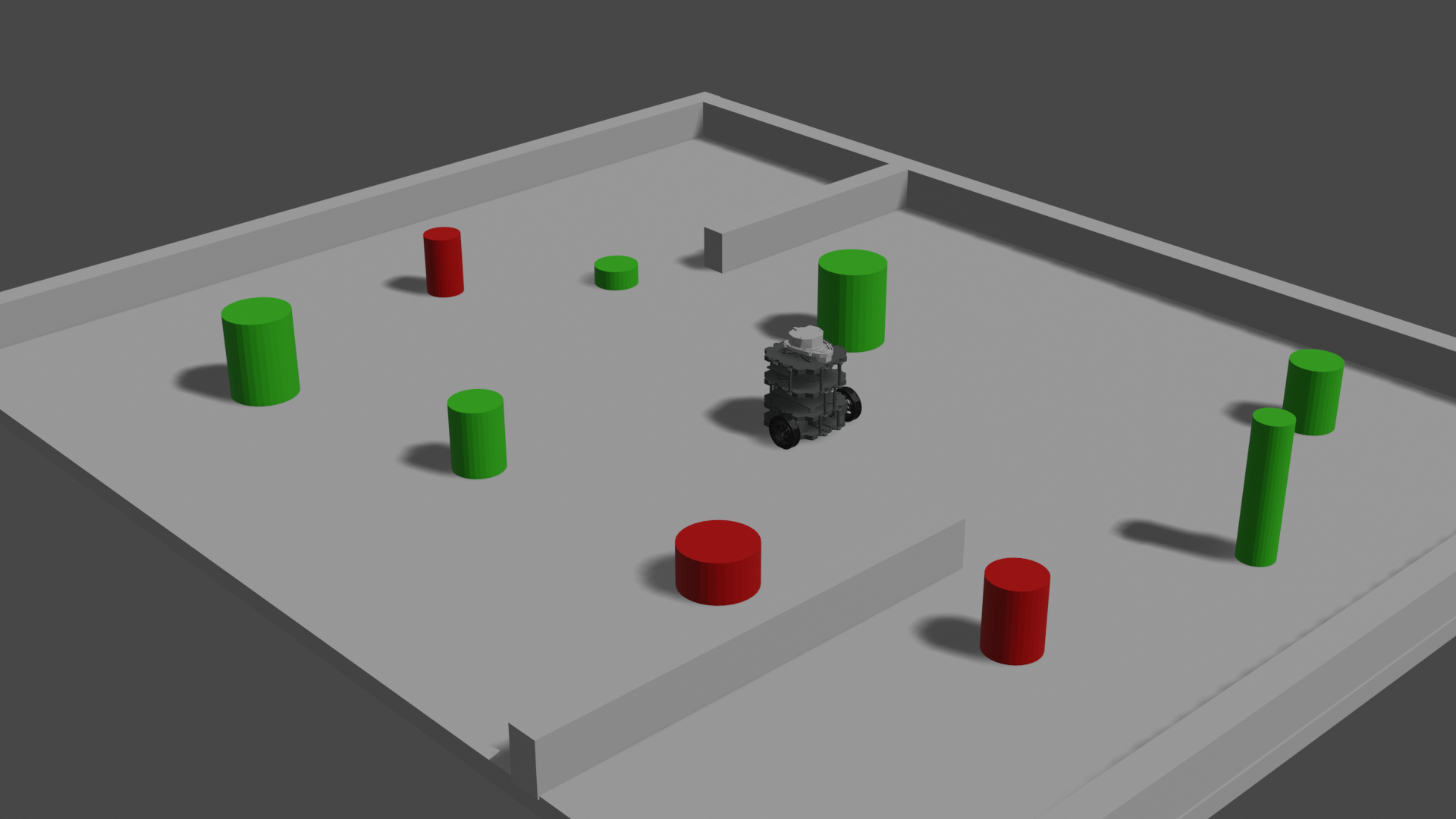}
\vspace{-0.6cm}
    \caption{Experimental setup used: at the center is a mobile robot, the green objects represent movable entities, at the opposite of the red ones. The room is closed.}
    \label{fig:setup}
\end{figure}

The experimental setup used in this article is presented in Figure \ref{fig:setup}.

A mobile robot, equipped with two controllable wheels is placed in a rectangular room. It is surrounded by multiple objects and possesses a LIDAR sensor, helping it to detect obstacles. It can navigate between them or learn to push them.
The robot starts with a limited prior knowledge: 
\begin{itemize}
    \item how many effectors it possesses, in this case its 2 wheels,
    \item a method to perform actions on its effectors. In our case, the method lists indexes of the effectors to activate and the action parameters $p \in [-1, 1]^2$ representing the intensity of the current to apply to each wheel.
    \item a list of the objects present in the room. We consider 9 cylinders in our setup, of various sizes and colors. The green objects are pushables whereas the red ones are not.
    \item a method to observe the properties of each of the objects (including the robot itself). The properties used in our experiment are: position, shape, radius, height, color, and the position relative to the robot. A pushable object requires more torque to be moved as its radius and height increase.
\end{itemize}

With this knowledge, the robot has to autonomously explore its environment and learn how to perform various tasks: moving itself, placing an object somewhere, pushing an object using another one. It also has to learn affordances corresponding to such tasks and to be able to recognise (or estimate) objects on which an affordance may apply or not. E.g. the pushability affordance cannot be applied to red objects, as those are fixed.

The robot explores the room in episodes of arbitrary length of 100 actions. At the beginning of each episode, the position and properties of the objects are initialised at random  values and the robot always starts at the center of the room. 

In real life operation, the robot can extract this information and properties from its sensors. It our case, the real life system can use an RGB-D sensor to segmentate objects and an variational autoencoder to extract properties from each of them. All the acquired data are used to fill in an environmental map, used in turn to provide data to the learning algorithm described in this article. To keep this system simple and avoid multiplying the source of errors, this article focuses only on the learning algorithm, with direct access to high-level data.

\subsection{Problem formalization}

Let us consider a robot interacting with its non-rewarding environment by performing sequences of motions of unbounded length in order to induce changes in its surroundings.

Each one of these motions is a primitive action 
described by a parametrised function with $N$ parameters : $a \in \mathcal{A} \subset \mathbb{R}^N$.
Each primitive action $a$
corresponds to a command that may be sent to one or several actuators of the robot.

Our robot can perform sequences of primitive actions. Such sequence may be of any length $n \in \mathbb{N}$, and described by $n$ successive primitive actions: $a = [a_1, \dots, a_n] \in \mathcal{A}^n$. Thus the primitive action space exploitable by the robot is a continuous space of infinite dimensionality $\mathcal{A}^{\mathbb{N}} \subset \mathbb{R}^{\mathbb{N}}$.

Each of the actions performed by the robot may have consequences on the environment, observable by the robot. We call such consequences \textit{observations} and note them $\omega \in \Omega \subset \mathbb{R}^M$. 

Each subspace of $\Omega$ is related to a given property of an object $o \in \mathcal{O}$ present in the environment (e.g. the position of an object). We consider this relation known to the robot, as such knowledge is required to build an affordance. This is a weak assumption as such information may be extracted from visual segmentation or by exploiting data from a semantic map. In our case, such data are directly given by the simulator itself.


\subsection{Formalization of our approach}


To learn how to interact with its environment, the robot learns models of relations between primitive actions $a \in \mathcal{A}$ and outcomes $\omega \in \Omega$ (relative observations before and after executing an action) obtained after performing this action within a given context $\tilde{\omega} \in \Omega$ (absolute state before executing the action, for more convenience we indicate with $\tilde{}$ the context spaces to differentiate them from outcome spaces).

For convenience, we define the controllable ensemble $\mathcal{C} = \mathcal{A} \cup \Omega_{controllable}$, regrouping both primitive actions $\in \mathcal{A}$ and observables that may be controlled ($\in \Omega_{controllable}$), i.e. that a model may be used to find one or a sequence of primitive actions to be performed in order to induce a value for the given observable. $\Omega_{controllable}$ is a subset of $\Omega$ and this set changes dynamically as the robot discovers new control models. 

More generally, the robot may learn models between controllables $c \in \mathcal{C}$ (not only primitive actions) and relative observations within a given context. Indeed, our robot may learn how to reach a goal observation value by first inducing a change in another observable of the environment. E.g.\ pushing an object can be performed after reaching the object.

To formalise affordances we use two elements. First we note an Affordance model $A(\mathcal{C}_i, \Omega_{j}, \tilde{\Omega_k})$, where $\mathcal{C}_i \subset \mathcal{C}$ is the input space of the affordance, $\Omega_{j} \subset \Omega$ is the output space and $\tilde{\Omega_k} \subset \Omega$ is its context space. And, secondly, to visually identify this affordance, we associate $A$ with a visual predictor $p_A$. It learns on $\Omega$ and indicates whether $A$ may be applied to an object $o$ in the scene or not, accordingly to its visual or physical properties. Moreover, to be able to learn how to use the affordance to complete tasks, $A$ possesses a forward model $M_A$ and an inverse model $L_A$. Both models learn the relationship between $\mathcal{C}_i$ and $\Omega_{j}$ knowing a context $\tilde{\Omega_k}$. The forward model is used to predict the observable consequences $\omega$ of a controllable $c_i$ in a given context $\tilde{\omega}$. Conversely, the inverse model is used to estimate a controllable $c_i$ to be performed in a given context $\tilde{\omega}$ to induce a goal observable state $\omega$ as a result of $c_i$. These models are trained on the data acquired by the robot all along its life and recorded in its dataset. Let us note $\mathcal{D}$ this dataset.


Each affordance $A$ can be seen as a basic skill, letting the robot perform a given simple task, e.g. reaching a position, placing an object somewhere.

Let us note $\mathcal{H}$ the ensemble of the affordances used by our robot. As our robot aimed to be adaptive, $\mathcal{H}$ varies along time.

\section{Algorithm}
For our robot to learn to associate a sequence of primitive actions $[a_1, ..., a_n]$ to desired consequences on multiple objects in its environment, our robot needs to learn which consequences $\omega$ can be observed and learn the control actions to realise these consequences.
For this learning problem, we propose an algorithm in this section. We first introduce its global architecture before detailing its key processes: how intrinsic motivation drives the exploration, how actions are executed and finally how affordance models are built and updated.

\subsection{Global architecture}

Our algorithm is based on the CHIME algorithm \cite{M2019}. Both are iterative and active learning algorithm that learn by episodes, but unlike CHIME our algorithm is designed to consider visual properties during its learning process.

\begin{figure}[h!]
    \centering
    \includegraphics[width=\linewidth]{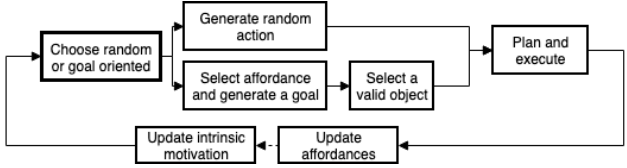}
    \vspace{-0.6cm}
    \caption{Abstract layout of a learning episode, beginning is on the left on the bold node.}
    \label{fig:architecture}
\end{figure}

The global layout of the algorithm architecture is presented in Figure \ref{fig:architecture} and the corresponding pseudo code can be seen on Algorithm \ref{alg:main}. At the beginning of the learning, the dataset $\mathcal{D}$ and the affordance hierarchy $\mathcal{H}$ are both empty: the robot autonomously collects data and creates affordances.

At each episode, the robot explores its environment by performing actions, observes the context and the outcomes obtained and processes the acquired data. One episode is composed of multiple iterations, and at each iteration one primitive action is performed.

Starting an episode, the robot decides either to explore a random action (l. \ref{alg:main:random}), or to use goal babbling to generate a goal to attain during the episode (l. \ref{alg:main:goal}). This decision is stochastic, based on a parameter $\sigma$, and it also depends on whether interesting goals may be generated or not. E.g. at the first episode, no data has been acquired yet and thus only a random action may be performed.

When choosing a random action, the robot generates a random controllable to be tested $c \in \mathcal{C}$ among all the controllable spaces (including the primitive actions). If required, this controllable is then converted to an executable primitive action, as only primitive actions may be performed by the robot effectors. This process is described in Section \ref{section:execution}.

When choosing to generate a goal, an affordance $A$ and a goal $\omega_g$ are selected, based on an interest metric detailed later in section \ref{section:im}. The robot next decides on which object this goal will be tested. Currently it just selects the closest object considered as valid by the affordance visual classifier $p_A$. The robot then uses its inverse models and its planning system to infer a sequence of controllables $c \in \mathcal{C}^n$ to be performed in order to reach $\omega_g$. Once again, these controllables are broken down into executable primitive actions, if required, using the same process as previously.

In both cases, the robot generates a sequence of primitive actions $a = [a_1, \dots, a_n] \in \mathcal{A}^n$ of length $n$. This corresponds to a random action or to a sequence of actions designed to reach a generated goal $\omega_g$. These actions are then executed by the robot (l. \ref{alg:main:execute}): for each sub primitive action $a_i$, the absolute value of each observable space is first recorded (corresponding to the context of the subaction), $a_i$ is then performed and the difference for each observable space (compared to before the execution) is retrieved.

After finishing an episode, the robot obtains a list of \\ $(a^i, \omega_1^i, \dots, \omega_k^i, \tilde{\omega}_1^{i}, \dots, \tilde{\omega}_k^{i})$ for each iteration. Where $i$ corresponds to the iteration index and $k$ to the number of subspaces of $\Omega$. These data are then stored in $\mathcal{D}$ (l. \ref{alg:main:store}). It is also processed and used to improve existing affordances (l. \ref{alg:main:update}), decide whether creating a new affordance is necessary or not, and update the intrinsic motivation system. These different processes are described in the following sub sections.

\setlength{\textfloatsep}{4pt}
\begin{algorithm}
    \caption{Algorithm layout
        \label{alg:main}}
    \begin{algorithmic}[1]
        \State{$i = 0$}
        \Loop
            \State{$\mathcal{D}_{episodic} = \emptyset$}
            \If{$\mathcal{H} \neq \emptyset$ \BoolAnd\ Random$() \leq \sigma$}
                \label{alg:main:goal}
                \State{$A =$ AffordanceSelection$(\mathcal{H})$} \label{alg:main:goal}
                \State{$\omega =$ GoalSelection$(A)$}
                \State{$\omega_g =$ ObjectSelection$(A, \omega)$}
                \State{$c =$ Plan$(\omega_g)$} \label{alg:main:goal-end}
            \Else
                \State{$\mathcal{C}_i =$ RandomControllableSpace$(\mathcal{C})$} \label{alg:main:random}
                \State{$c_r =$ RandomValue$(\mathcal{C}_i)$}
                \State{$c = [c_r]$} \label{alg:main:random-end}
            \EndIf
            \State{$a =$ TransformToPrimitive$(c)$}
            \For{$a_k \in a$} \label{alg:main:sk}
                \State{$\omega_{before} =$ GetObservations$(\Omega)$}
                \State{$a_i = [c_k] \ \textbf{if}\ c_k \in \mathcal{A} \ \textbf{else}$ TransformToPrimitive$(c_k)$} \label{alg:main:primitive}
                \State{Execute$(a_i)$} \label{alg:main:execute}
                \State{$\omega_{after} =$ GetObservations$(\Omega)$}
                \State{$\omega_i = \omega_{after} - \omega_{before}$} \label{alg:main:relative}
                \State{$\tilde{\omega_i} = \omega_{before}$} \label{alg:main:context}
                \State{$\mathcal{D}_{episode} \leftarrow (a_i, \omega_i, \tilde{\omega_i})$} \label{alg:main:store-end}
                \State{$i \mathrel{+}= 1$}
            \EndFor
            \State{UpdateInterestMaps$(\mathcal{D}, \mathcal{D}_{episodic})$}
            \State{UpdateAffordances$(\mathcal{D}, \mathcal{D}_{episodic})$} \label{alg:main:update}
            \State{$\mathcal{D} \leftarrow \mathcal{D}_{episodic}$} \label{alg:main:store}
        \EndLoop
    \end{algorithmic}
\end{algorithm}

\subsection{Intrinsic motivation} \label{section:im}

This algorithm uses intrinsic motivation to guide its exploration. It is based on the CHIME algorithm \cite{M2019}, itself inspired by the SAGG-RIAC algorithm \cite{baranes_oudeyer2009_riac}.

For each affordance $A(\mathcal{C}_A, \Omega_{A}, \tilde{\Omega_A})$, the system creates an interest map: a partition of $\Omega_{A}$ that is constructed incrementally based on progress measures as described in \cite{baranes_oudeyer2009_riac}. The goal of this process is to divide $\Omega_{A}$ into regions and attribute a value of interest to each region. This interest corresponds to a monitoring of how much exploring this region may improve the robot knowledge in the future.

This measure is linked to a notion of \textit{competence}. In our case, we define the competence of an affordance $A$ near a goal $\omega \in \Omega_{A}$ as $mean(\omega_e - \omega_r)$ for the $k$ last outcomes near $\omega$. $\omega_e$ corresponds to an outcome goal estimated by the algorithm for a given controllable $c$ and $\omega_r$ to the effective outcome reached during the exploration.


The derivative of this \textit{competence} is used to define a \textit{learning progress}: how much an affordance model has been improved. And the \textit{interest} value of a region then corresponds to the mean of the last $n$ \textit{learning progresses} in this region.

More details about this process and the region splitting mechanism may be found in \cite{M2019} and in \cite{baranes_oudeyer2009_riac}.

\subsection{Action and controllable execution \label{section:execution}}

To perform sequence of controllables $c$, our algorithm uses the same system as CHIME.
For each element $c_i$ of $c$:
\begin{itemize}
    \item if the sub-controllable to be performed $c_i$ is a primitive action, it is directly sent to the effectors and executed without any pre-processing
    \item in the other case, if $c_i$ is not a primitive action, it corresponds to an observable the robot wants to induce within its environment, i.e. $c_i \notin \mathcal{A}$ but $c_i \in \Omega_{controllable}$. Then it cannot be directly executed by the effectors of the robot and it needs to be broken down into primitive actions beforehand. An affordance $A(\mathcal{C}_A, \Omega_{A}, \tilde{\Omega_A})$ is then selected (with $c_i \in \Omega_{A}$) and its inverse model is applied onto $c_i$ in order to obtain a lower level controllable $b_i \in \mathcal{C}_A$. If $c_i$ is difficult to reach using only one lower level controllable, a planning phase is used to build a sequence of element of $\mathcal{C}_A$ in order to reach $c_i$ when executed. Once again for each element of this newly created sequence, if it is not primitive the same mechanism is applied recursively on it until having only primitive actions.
\end{itemize}

At the end of this mechanism, we obtain a list $b \in \mathcal{A}^\mathbb{N}$ composed only of primitive actions that can be executed directly.

Additional information can be found in \cite{M2019}.

\subsection{Affordance addition and update}

The CHIME algorithm has been designed to autonomously learn model of data.
We diverge from it to autonomously learn affordances instead. In this section we present how affordances are added to $\mathcal{H}$ and updated.

At each episode, the robot has to decide multiple elements: whether a new affordance must be added or if existing affordances are enough; how to train the visual classifiers of affordances and if affordances need to be updated.

To answer those questions, the robot follows the procedure presented on Algorithm \ref{alg:aff}.

At the end of each episode, subspaces of $\Omega$ for which non null relative outcomes $\omega$ has been observed are listed. Then the robot randomly picks a space among this list and verifies if it matches an existing affordance. A space matches an affordance if adding the data from this space to the training set of the affordance does not reduce its competence. If not matching, it tries to add context spaces to the affordance or then tries to create a new affordance. The predictor $p_A$ is afterwards trained on the acquired data (positive or negative).

The predictor $p_A$ used in our system is a binary neural network composed of 3 fully connected layers using as input all the properties of the object $o$ currently considered. It is trained using action replay on balanced data (objects on which the $A$ is applicable and the others).


\setlength{\textfloatsep}{4pt}
\begin{algorithm}
    \caption{Autonomous affordances adaptation
        \label{alg:aff}}
    \begin{flushleft}
        \hspace*{\algorithmicindent} \textbf{Input:} $a$ the actions performed during the episode, \\
        $\omega$ the observations at the beginning of each iteration of the episode.
    \end{flushleft}
    \begin{algorithmic}[1]
        \State{$Spaces = $ SelectSpaces$(\Omega, \omega)$}
        \MRepeat{\ $k$ times}
            \State{$S = $ PickSpace$(Spaces)$}
            \For{$A \in \mathcal{H}$} \label{alg:aff:sk}
                \State{$matched = False$}
                \If{Matches$(A, a, \omega_S)$}
                    \State{$matched = True$}
                    \State{Add $(a, \omega_S)$ to the model training dataset of $A$}
                    \State{TrainVisualClassifier$(A, \omega_S, True)$}
                \Else
                    \MRepeat{\ $k'$ times}
                        \State{$S_{context}' =$ PickSpace$(\Omega)$}
                        \State{$NewA =$ Copy$(A)$}
                        \State{$ContextSpace_{NewA} = ContextSpace_{NewA} \cup S_{context}'$}
                        \If{Competence$(NewA) \geq \tau_{modification}$}
                            \State{$A \leftarrow NewA$}
                            \State{$matched = True$}
                            \State{\textbf{break}}
                        \EndIf
                    \EndRepeat
                \EndIf
            \EndFor
            \State{$p_A \gets$ TrainVisualClassifier$(A, \omega_S, matched)$}
            \If{$matched$}
                \State{Add $a, \omega_S$ to the model training dataset of $A$}
            \Else
                \State{$NewA =$ Affordance$(a, S, \emptyset)$}
                \If{Competence$(NewA) \geq \tau_{creation}$}
                    \State{$\mathcal{H} \leftarrow NewA$}
                    \State{$p_{NewA} \gets$ TrainVisualClassifier$(NewA, \omega_S, True)$}
                \EndIf
            \EndIf
        \EndRepeat
    \end{algorithmic}
\end{algorithm}

\section{Performance of our algorithm}

We used our experimental setup to perform three series of tests:
\begin{itemize}
    \item firstly, evaluating our system itself, which affordances are created, and when;
    \item then, comparing the task performance of our system compared to CHIME;
    \item finally, comparing our system to other reinforcement learning approaches.
\end{itemize}

\subsection{Evaluation method}

To measure the performance of our (or other) algorithm at completing tasks, we define an evaluation metric as follows: for each task, we pre-define a list of points (robot position or object position) to be reached. Then, during evaluation, the system attempts to reach each point in the simulator and $1 -$ the mean error at reaching those points is defined as the evaluation of this task.

\subsection{Affordances learning}

In our first test, we let the robot explore the environment presented in \ref{section:setup}. This environment is simulated in \textit{python} using a 2D physics engine named \textit{pymunk}.

We perform 10 runs, letting the robot autonomous during 4000 iterations and we report the mean results.

\begin{figure}[h!]
    \centering
    \includegraphics[width=1.\linewidth]{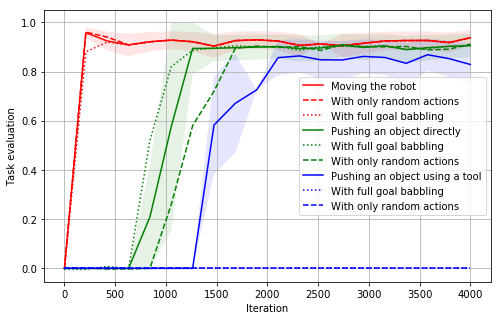}
    \vspace{-0.8cm}
    \caption{Evaluation during the training for three affordances: moving the robot itself $A_0$, pushing an object $A_1$ and pushing an object using another one $A_2$. Evaluation is done every 200 iterations between 200 and 2000. Thus, affordance $A_1$ is created between 600 and 800. This also shows mean evaluation value when using only random actions or only goal babbling when possible (standard deviation for those is not displayed for clarity). For $A_3$, the goal babbling or random only versions does not manage to create the affordance.}
    \label{fig:competence}
\end{figure}

At the end of its exploration, we observe the affordances created and their evaluation, as presented in Figure \ref{fig:competence}. The robot has successfully discovered multiple affordances, we count 12 at the end for the majority of runs. Among them, 3 where expected:
\begin{itemize}
    \item $A_1$: moving the robot itself,
    \item $A_2$: pushing an object by moving the robot and
    \item $A_3$: pushing an object using another object. 
\end{itemize}
The other affordances discovered are unintended, but still valid: they correspond to unexpected correlations the robot has found between various spaces.
In our analysis we focus on the first 3 affordances mentioned above.

Even in this simple environment, the algorithm has managed to create a hierarchy of interrelated skills: $A_2$ depending on $A_1$ to be completed, itself depending on $A_0$.

More than just the final number of affordances, it is interesting to observe the creations, deletions and updates of affordances all along the exploration.

\begin{figure}[h!]
    \centering
    \includegraphics[width=0.65\linewidth]{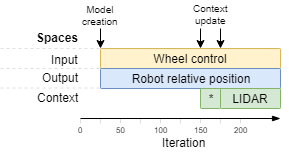}
    $A_0$
    \includegraphics[width=0.85\linewidth]{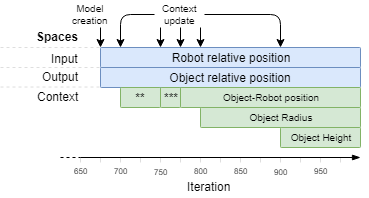}
    $A_1$
    \vspace{-0.6cm}
    \caption{Temporal evolution of affordances $A_0$ (top) and $A_1$ (bottom) during the learning process. Please note that the iteration axis is not the same for $A_0$ and $A_1$. Colors are not related to the competence graph: yellow spaces are part of $\mathcal{A}$, blue and green ones of $\Omega$: blue ones are using relative data while green ones absolute data.
    \\
    \\
    *    : relative position between an object and the robot \\
    **   : robot absolute position\\
    ***  : LIDAR data}
    \label{fig:affordances}
\end{figure}

Concerning the affordance $A_0$, we can see in Figure \ref{fig:affordances} (top) that the affordance is created since iteration t=25. At the moment of the discovery of this affordance, the model created by the robot does not take as input any context space. As no walls or obstacles have been encountered yet the robot thinks the movement of the robot only depends on its wheels speed. At iteration t=150 the affordance is updated and the relative position between an object and the robot is added as context space. A wrong assumption but coherent with the data acquired so far. Then quickly, at iteration 175, this context space is replaced with the robot LIDAR space and kept as such until the end. No physical properties is used here as a context space, this is due to the fact that the robot is the only object using this affordance.


The results of $A_1$ in Figure \ref{fig:affordances} (bottom) show that this affordance is created much later than $A_0$. This is explained by the fact that the robot has first to collide with an object to discover how to push objects directly with its body. The first occurrence of such collision was around t=500 iterations on average. Here again, the context space of the affordance has evolved during the exploration and has finally converged to the relative position between the object and the robot. At the difference of $A_0$, 2 physical properties are here added as context spaces of this affordance: the radius and the height of the object at hand. As the pushability of each object depends on these two physical properties it is normal to see them appear here, and this confirm that our algorithm has well captured the dependency to such properties.

Once $A_1$ has been created, its visual classifier $p_{A1}$ is also created and trained to identify to which object $A_1$ may be applied or not. At the end of the 4000 iterations, we use $p_{A1}$ to check its prediction for each object in the room including the robot itself: it is positive for all the green objects and negative for the robot. This is expected as the robot cannot push itself neither it can push fixed red objects. Hence, our algorithm has successfully managed to construct both a model affordance and the corresponding visual classifier.

For $A_0$, $A_1$ and $A_2$, the affordances are created directly as soon as collected data permit it. This behaviour is desired and due to a low affordance creation threshold $\tau_{addition}$. This favours exploration of newly discovered spaces and regions: indeed, with a low threshold value, affordances are easily created and a goal may be generated to explore them. If the exploration then points out that it is a false positive, that affordance is destroyed. On the contrary if the exploration confirms it as a valid affordance, active learning continues to gradually collect new data to increase the robot's competence for this affordance.

\subsection{Random and Goal Babbling impact}

To further analyse our algorithm we decided to test two extreme situations: one with only random action exploration; and another one using only goal babbling whenever possible.

The first case favours novelty and discovery: the rate of affordance addition is high, but the exploration and the mastering of the already discovered affordance is delayed. In Figure \ref{fig:competence} we can see that the competence curve for $A_1$ requires more time to converge than in the previous test.

At the opposite, using only goal babbling whenever available, the number of affordances discovered is greatly reduced, and focused at the beginning of the exploration. In this configuration, $A_2$ is discovered later compared to the previous configuration.




\section{Comparison with other approaches}

We compare our approach to baselines belonging to two different families: firstly to reinforcement learning algorithms on similar setups. Secondly, we compare it to affordance learning algorithms. But to our knowledge such methods do not focus on mobile robot and are thus evaluated on experimental setups significantly different from ours.

\subsection{Reinforcement learning}

As we want to compare our algorithm to existing ones on the same setup, we choose to use classical reinforcement learning algorithms such as Q-Learning, DQN (Deep Q-Network) and Actor Critic in our experimental setup. As they are not designed for multi-task learning and require an extrinsic reward, some setup modifications have been made to enable these algorithms to learn in our setup: we limited the experiment to one object at a time (except for $A_2$) and added a reward function to provide a feedback. Unlike our method, where the exploration is self-guided, the desired behaviours or tasks to be completed with these algorithms must be explicited through the reward function. We test these algorithms on 3 increasingly difficult tasks: moving the robot, pushing an object directly and then by using another object as a tool.
To match the general aspect of our algorithm, we use Universal Value Function Approximators \cite{Schaul2015} for these three algorithms in order to learn how to reach various goals.
We use 2 different kinds of reward function for each setup:


\begin{table}[h!]
\begin{tabular}{|l|c|c|l|}
\hline
\textbf{Version}                                                              & \multicolumn{1}{l|}{\textbf{Reaching goal} } & \begin{tabular}[c]{@{}c@{}}\textbf{Pushing/going}\\ \textbf{in the right direction}\end{tabular} & \textbf{Else}                    \\ \hline
\begin{tabular}[c]{@{}l@{}}Non-guided\\ (sparse reward)\end{tabular} & +1000                              & \multicolumn{2}{c|}{-5}                                                                                  \\ \hline
Guided                                                               & +1000                              & max +20                                                                        & \multicolumn{1}{c|}{-5} \\ \hline
\end{tabular}

\caption{Reward functions used by the comparative setup}
    \vspace{-0.8cm}
\end{table}

In addition to these algorithms, we also compare ours to CURIOUS, a reinforcement learning algorithm using intrinsic motivation for exploration. We base its reward on the non-guided version.

When required, $\Omega$ has been discretised uniformly. Actions have also been discretised into 4 when needed: forward, backward, turning left, turning right. When reaching the zone or after 1000 iterations the episode ends, the setup is reset and the robot is randomly placed inside the room.

\begin{figure}[h!]
    \centering
    \includegraphics[width=\linewidth]{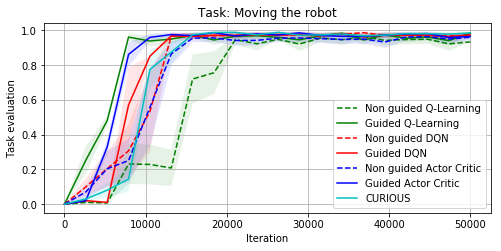}
    \\
    \includegraphics[width=\linewidth]{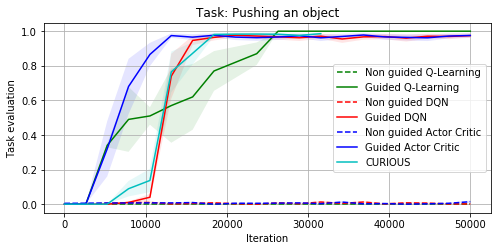}
    \\
    \includegraphics[width=\linewidth]{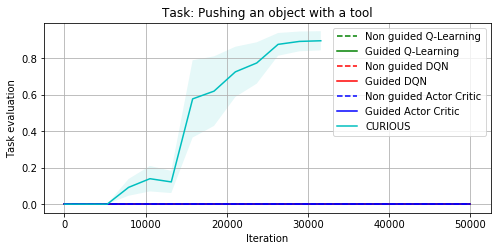}
    \vspace{-0.8cm}
    \caption{Evaluation of Q-Learning, DQN, Actor Crictic and CURIOUS applied to three tasks: moving the robot, pushing an object and pushing it using another object as a tool. The standard deviation is displayed in transparent.}
    \label{fig:qlearning}
    \vspace{-0.4cm}
\end{figure}

We perform 10 runs over 50000 iterations for each task, reward function and algorithm and report the result in Figure \ref{fig:qlearning}.

For the first task (top), we can see that all the algorithms succeed in 10000 to 20000 iterations. With our algorithm, moving the robot is mastered as soon as 250 iterations. The difference mainly comes from the use of planning in our case. It lets the robot reach distant spots even with such a few exploration done.

For the second task (middle), only the guided version are successful, requiring between 12500 and 26000 iterations to be learned. The non-guided versions fail because of the combinatorial explosion of all the states involved and the difficulty to reach the final goal. In our case this task is learned within 1000 iterations.

For the  most complex task (bottom), only CURIOUS and our algorithm manage to succeed, CURIOUS reaches a competence of 0.96 using 32000 iterations. The other algorithms, even using the guided rewards, fail due to the complexity of the task at hand. Our algorithm only requires 2100 iterations to reach the final level of competence of CURIOUS.

For all the examples above, as we use UVFA, the Q-Learning is highly dependent to the number of goal it has to explore (as each goal corresponds to a different state). Thus, this adds another prior (in addition to the reward function) that is not required with our algorithm.

\subsection{Affordance learning}

As the majority of works in affordance learning uses robot arms to manipulate objects and not mobile robot, experimental setups are difficult to compare. Thus, we decide to provide a qualitative comparison between our approach and existing affordance learning ones. We analyse the learning process reported for the subsequent affordances in these different setups.

In \cite{montesano_lopes2008_affordance_learning_coordination_imitation}, the system extracts pre-programmed controllables from the considered objects, like in our algorithm, then discretises them and clusterises them. It then builds a dependency graph that encompasses visual controllables, performed action and the action context. In our case, the information contained in this graph are all included in our models and visual classifiers. Thus, our system is capable to build the same affordances. Conversely, the pre-programmed actions in \cite{montesano_lopes2008_affordance_learning_coordination_imitation} are in our case autonomously learned by the robot, requiring less prior information and adding more flexibility. In both cases, the temporal aspect of sequences of actions is not learned, but in our algorithm, the planning layer automatically creates successive sequences, based on the models learned.

On the contrary, in \cite{Ugur2016}, the system builds a hierarchy of affordances like in our proposition. This time intrinsic motivation is used to select which action to execute within a finite set of pre-defined low level actions. Whereas in our system, the robot manages to learn primitive actions in a continuous space, and is capable to use sequences of actions by chaining primitive actions.

\section{Conclusion}
For affordances learning, we have presented an algorithm combining the affordances concept and intrinsic motivation exploration. It allows a robot to autonomously discover unknown affordances and learn actions to exploit them. The learning is based on active learning to collect data through new interactions with the environment, guided by the heuristics of intrinsic motivation; Once learned, these affordance control models are used to plan complex tasks with known or unknown objects, by using their physical properties to decide whether or not a learned affordance may be applied.

Our main contribution in this article is to propose a learning algorithm for multiple objects based on physical properties so as to generalise to new objects. We have shown that it can discover in a developmental manner non-predefined affordances from the easiest to the most complex ones, and can use unbounded sequences of learned actions to complete complex tasks.
We have compared our algorithm to others to outline two main properties : the hierarchical and developmental learning process, as well as the capacity to use sequences of actions to adapt to the complexity of the task at hand.

This algorithm broadly relies on the concept of embodiment and is strongly inspired by human development from this point of view; for both the affordance aspect and the intrinsic motivation one.

In future works we want to deepen the comparison with existing methods by considering similar setups, and thus applying our algorithm onto robotic arms. Also, we aim for a more complete system, including a mechanism for visual feature extraction in order to provide inputs for our algorithm.


\bibliographystyle{ACM-Reference-Format}
\bibliography{lectures}

\appendix

\end{document}